# Generator evaluator-selector net for panoptic image segmentation and splitting unfamiliar objects into parts

Sagi Eppel[1,2], Alán Aspuru-Guzik[2,3,4,5]

[1]Vector Institute for Artificial Intelligence, Toronto, Canada.  [2]Department of Chemistry, University of Toronto, Canada. [3]Department of Computer Science, University of Toronto, Canada. [4]Department of Chemistry and Chemical Biology, Harvard University, Cambridge, USA. [5]Canadian Institute for Advanced Research (CIFAR) Senior Fellow, Toronto, Canada

## Abstract

In machine learning and other fields, suggesting a good solution to a problem is usually a harder task than evaluating the quality of such a solution. This asymmetry is the basis for a large number of selection oriented methods that use a generator system to guess a set of solutions and an evaluator system to rank and select the best solutions. This work examines the use of this approach to the problem of panoptic image segmentation and class agnostic parts segmentation. The generator/evaluator approach for this case consists of two independent convolutional neural nets: a generator net that suggests variety segments corresponding to objects, stuff and parts regions in the image, and an evaluator net that chooses the best segments to be merged into the segmentation map. The result is a trial and error evolutionary approach in which a generator that guesses segments with low average accuracy, but with wide variability, can still produce good results when coupled with an accurate evaluator. The generator consists of a Pointer net that receives an image and a point in the image, and predicts the region of the segment containing the point. Generating and evaluating each segment separately is essential in this case since it demands exponentially fewer guesses compared to a system that guesses and evaluates the full segmentation map in each try. The classification of the selected segments is done by an independent region-specific classification net. This allows the segmentation to be class agnostic and hence, capable of segmenting unfamiliar categories that were not part of the training set. The method was examined on the COCO Panoptic segmentation benchmark and gave results comparable to those of the basic semantic segmentation and Mask-RCNN methods. In addition, the system was used for the task of splitting objects of unseen classes (that did not appear in the training set) into parts.

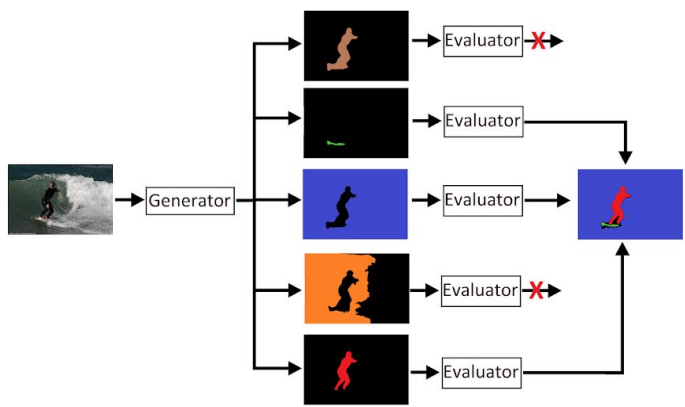 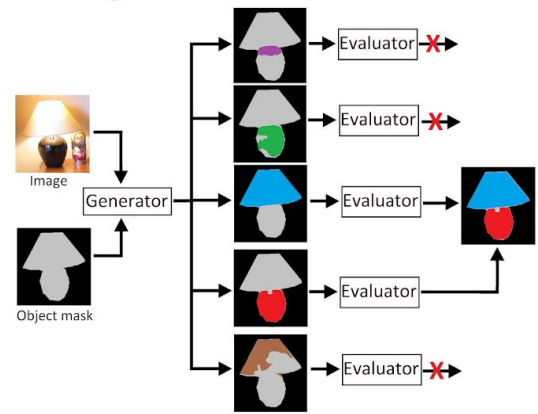

# 1. Introduction

For many problems in machine learning and other fields, the task of generating a good solution is significantly harder than evaluating the quality of such a solution [1-3]. Solving problems in these kinds of systems is often done by combining two processes: a generator process that guesses solutions to the problem and a selector process that filters low-quality solutions and picks high-quality solutions [1-3]. Hence, an inaccurate generator can suggest random solutions with wide variability in quality, while the accurate selector system grades and selects the best solutions. This work will examine the use of the generator evaluator selector (*GES*) approach for the tasks of splitting unfamiliar objects into parts and panoptic image segmentation [4]. This will be done by combining a generator net[5] that guesses various segments corresponding to object, stuff and parts regions in the image and an independent evaluator net that ranks and selects the best segments to be used in the final segmentation map (Figure 1).

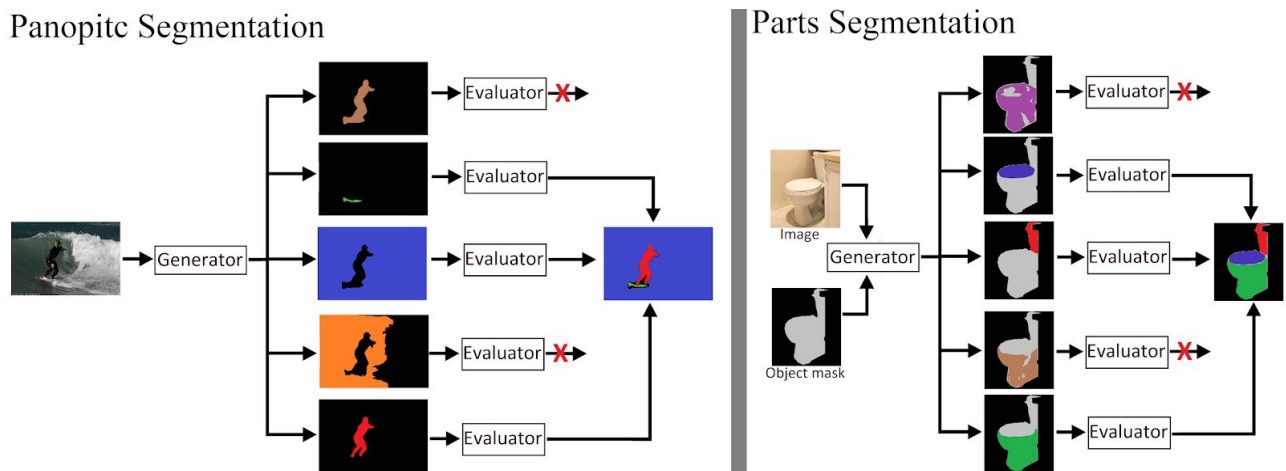

**Figure 1: Image segmentation using the generator evaluator/selector method. Panoptic segmentation: the panoptic generator guesses various segments corresponding to object instances (person, board) and non-object regions with distinct classes (sea, sky). The evaluator system grades and selects the best segments to be merged into a final segmentation map. Parts segmentation: the parts generator receives the object region and suggests object parts, the evaluator net rank, and filter segments that don't match real parts.**

The task of splitting an unfamiliar object into parts involves taking the region of the object in the image and segmenting its parts (Figure 1), for unseen object classes that did not appear in the training set. As far as we are aware, this is the first time such a problem has been explored with neural nets. Panoptic segmentation [4] is a combination of the two main image segmentation tasks: semantic segmentation which involves splitting the image into segments that cover different semantic classes, such as sky and sea (Figure 2), and instance segmentation, which involves splitting the image into segments corresponding to individual objects, such as individual people in a group (Figure 2). To date, the dominant approach for solving this task is using a *fully convolutional net* (FCN) [6] to assign a class for each pixel in the image, combined with the *Mask R-CNN* method [7] which finds bounding boxes for each object in the image and then assigns a mask for the object in each bounding box.

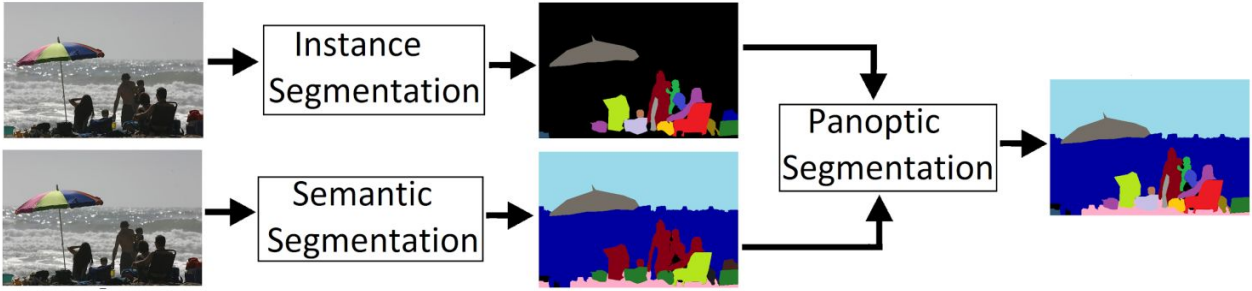

Figure 2: Instance segmentation involves splitting the image into regions corresponding to individual objects instances (people, chairs). Semantic segmentation involves finding the class of each pixel in the image (ocean, person). Panoptic segmentation is the combination of the two.

One problem with applying the generator selector approach using these methods is that they generate a single solution instead of a distribution of solutions. Selection-oriented systems are only useful when it is possible to generate several different solutions with a wide distribution of qualities [1-3]. Another problem with applying a selection-oriented approach with existing methods is the fact that they generate the full segmentation map instead of individual segments. For the generative selective process, searching for a solution by generating and evaluating each part of the solution separately will take exponentially fewer steps compared to guessing and evaluating the full solution in each try, as we will describe in Section 3.3.

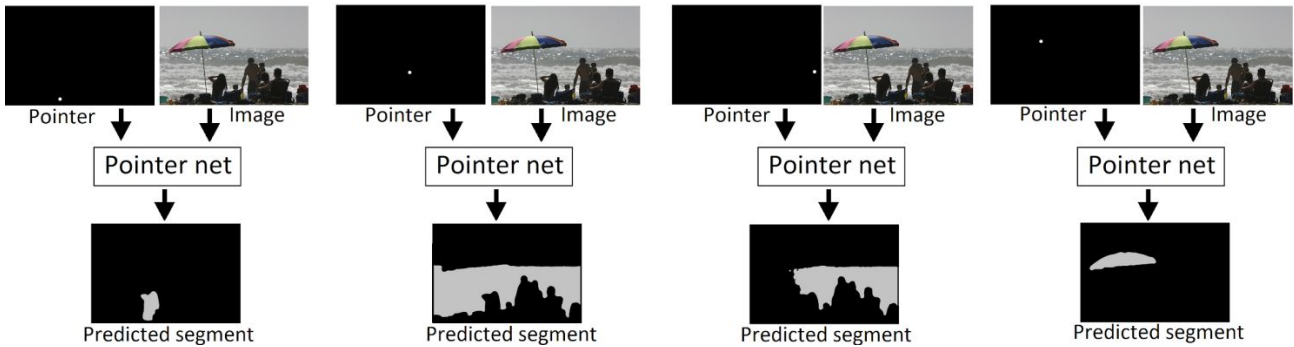

Figure 3: Pointer net as a segment generator. Given an image and a pointer point, the net predicts the segment containing the point. Selecting input points in different locations will lead to different output segments, even for two different points located inside the same segment.

To address these problems, we replaced the standard Mask R-CNN and semantic segmentation methods with a single pointer segmentation net (Figure 3). Pointer net [8] is an FCN that, given an image and a point in the image, finds the segment that contains the input point (Figure 3). The variability in the predicted segments emerges from the location of the input pointer point. Pointer net will produce different predictions even for different points located in the same segment (Figure 3). The evaluator system consists of a simple CNN that receives an image and a segment mask and returns the segment grade. This grade is basically the estimation of the match quality between the input segment and an actual segment in the image. Both the generator and evaluator are category independent and can be used to segment regions with unfamiliar classes that did not appear in the training set. To classify each segment, a region-specific classification neural network is used. This net receives the segment mask and the image, and returns the segment category [9, 10]. The combined method was evaluated on the COCO Panoptic benchmark [4, 11] for panoptic segmentation, and the ADE20K[40] and Pascal Parts[39] for parts segmentation.

## 2. Related work

Generative and selective systems

Selection-oriented systems that consist of a generator process that generates a variety of random products, and a selector process that evaluates and filters the products according to some selection method, appear in a wide range of fields [1-3]. In the field of machine learning and artificial intelligence, this type of system has been used in a wide range of methods such as genetic algorithms [12, 13, 14] and game playing reinforcement learning methods such as AlphaGo, in which one system generates suggestions for the game's next move, while another independent system ranks these moves [15, 16]. Another approach is generative adversarial nets (GAN), which uses two nets, one that generates a solution to some problem, and a second that tries to discriminate between the generated solution and the real solution [17, 18]. Unlike GES, the GAN works in a competitive manner and mainly as a training tool. Another related approach is adding a confidence score to the net prediction, which can be used to filter inaccurate predictions; this is usually done by adding a confidence head to the net or using the prediction consistency [19-23].

Panoptic segmentation

The panoptic segmentation task [4] consists of the recognition and segmentation of every individual object in the image as well as assigning a category for every pixel in the image. To date, all methods that were ranked on this challenge were based on combining two different heads for semantic segmentation and object recognition. Semantic segmentation is achieved using FCN [6] that assigns a class to each pixel in the image. Object recognition is mostly performed using some variation on the Mask R-CNN [7] method, which finds a bounding box for each object in the image and the segment mask of the object in the box. A number of improvements have been suggested to this method, including attention mechanism [24] and occlusion heads to unify the semantic and instance maps [25-30]. Mask R-CNN, as well as most object detection methods, contain a ranking mechanism for objectness of the proposed bounding box, which allows them to filter bounding boxes with low scores [31, 32, 33].

Open set and class agnostic segmentation

Most of the above methods are limited to either objects or non-object (stuff) regions and tend to be class-specific. Open set segmentation [34, 35] involves segmenting objects and stuff corresponding to unseen classes that did not appear in the training stage. This can be achieved by predicting the edges of objects and stuff in the image and using these boundaries to split the image into segments [34, 36, 37]. Another approach for open-set segmentation is the pointer net method, which receives a point and an image and returns the mask of the segment containing the point [8, 38]. By training on a large number of segments corresponding to different objects and non-objects (stuff), this net manages to achieve class agnostic segmentation of both objects and stuff [8]. This approach was significantly improved by adding a preselection net, which predicts which pointer points on the image are likely to produce good segments [38].

Parts segmentation

Two data sets exist for parts segmentation: The Pascal Parts [39] and the ADE20K [40, 41]. The Pascal Parts dataset contains only 20 categories and 10,000 images. The ADE20K contains 120 categories of objects and stuff with parts. While standard semantic and instance segmentation can segment parts by treating them as different classes or instances [42, 43], none of these methods were used to predict parts of unseen objects categories that were not used in the training stage.

# 3. Generator evaluator selector method

## 3.1. Panoptic generator evaluator method

A schematic for the full modular system for panoptic segmentation is shown in Figure 4. The method consists of four independent networks combined into one modular structure. The first step is generating several different segments using the pointer net as the generator. The segments generated by this net are restricted to a given region of interest (ROI) which covers the unsegmented image region. The generated segments are then ranked by the evaluator net. This net assigned each segment a score that estimates how well it corresponds to a real segment in the image. The segments which receive the highest scores and are consistent with each other are selected, while low-ranking segments are filtered out. The selected segments are then polished using the refinement net. Each of the selected segments is then classified using the classifier net. Finally, the selected segments are stitched into the segmentation map (Figure 4). The segmentation map is passed to the next cycle which repeats the process in the remaining unsegmented image regions. The process is repeated until either the full image has been segmented or the quality assigned to all of the predicted segments by the evaluator drops below some threshold.

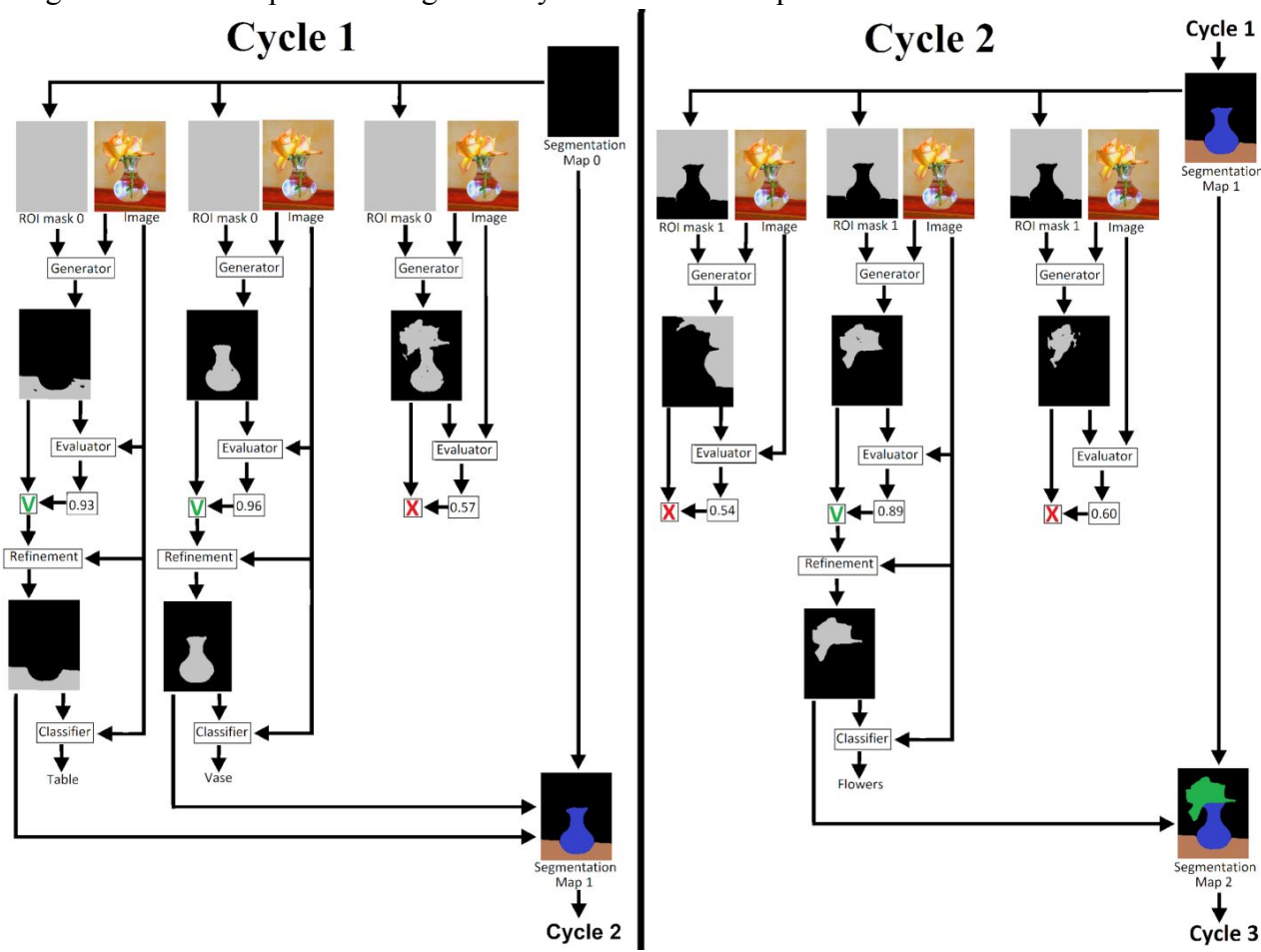

**Figure 4:** Modular generator/selector segmentation system. ROI mask is generated using the unsegmented regions of the image. The generator net guesses various segments within the ROI region. The suggested segments are ranked using the evaluator net. Segments with low scores are filtered out. Segments with high scores are refined using the refinement net and classified using the classifier net. The selected segments are added to the segmentation map. The updated segmentation map is moved to the next cycle.

## 3.2. Generator-evaluator method for parts segmentation

Applying the system for the task of splitting objects into parts was done using only the generator and evaluator nets (Refinement or classification were not used). The Parts generator is again a pointer net that receives a point in the image and a mask corresponding to the object region. The pointer net output is a mask corresponding to the part containing the input point (Figure 5b). The evaluator, in this case, is a net that receives the image, the predicted segment, and the mask of the object region. The evaluator outputs a score that corresponds to how well the input segment matches a real part in the image (Figure 5e). Note that the main difference between the Panoptic and part segmentation is that the ROI mask is replaced by the object mask and used as an input for both the generator and evaluator (Figure 5). The full segmentation of objects into parts was done by giving the generator 100 random points and evaluating the generated segments using the evaluator. The predicted segment was then added to the annotation mask one by one in the order of the score given to them by the evaluator. Predicted masks that have an overlap of more than 50% with previously added masks were ignored. If the overlap is less than 50%, the area of overlap was removed from the added mask.

### Segment generation using Pointer net.

Pointer net [8,38] acts as the segment generator, which creates proposals for different segments in the image (Figure 5.a,b). Pointer net receives an image and a point within this image. The net predicts the mask of the segment that contains the input point (Figures 3,5.a). In this work, the pointer point location is chosen randomly within an ROI region in the image (Figure 5.a). The net will predict different segments for different input points, even if the points are located within the same segment (Figure 3). While this feature was not planned, it allows the pointer net to act as a random segment generator with the ability to generate a large variability of segments by selecting random input points. Another input of the pointer net is a region of interest (ROI) mask, which restricts the region of the predicted segments (Figure 5a). The generated output segment region will be confined to the ROI mask. For part segmentation, the ROI mask is the region of the object containing the part (Figure 5b). While for panoptic segmentation, the ROI mask is the unsegmented region of the image.

### Evaluator net.

The evaluator net is used to check and rank the generated segments. The ranking is done according to how well the input segment fits the best matching real segments in the image. The evaluator net is a simple convolutional net that receives an image and a generated segment mask (Figure 5d-f). The evaluator net predicts the *intersection over union* (IOU) between the input segment and the closest real segment in the image. For the parts evaluator (Figure 5f), the net also receives the mask with the region of the object containing the part as an input.

### Refinement net.

The refinement net is used to polish the boundaries of the generated segment. The net receives the image and an imperfect segment mask. The net output is a refined version of the input segment (Figure 5c). This approach has been examined in several previous studies [45, 46].

Classifier net.

Determining the segment category is done using a region-specific classification net. The net receives the image and a segment mask. The net predicts the category of the input segment (Figure 5f). This approach has been explored in previous works [9, 10].
.

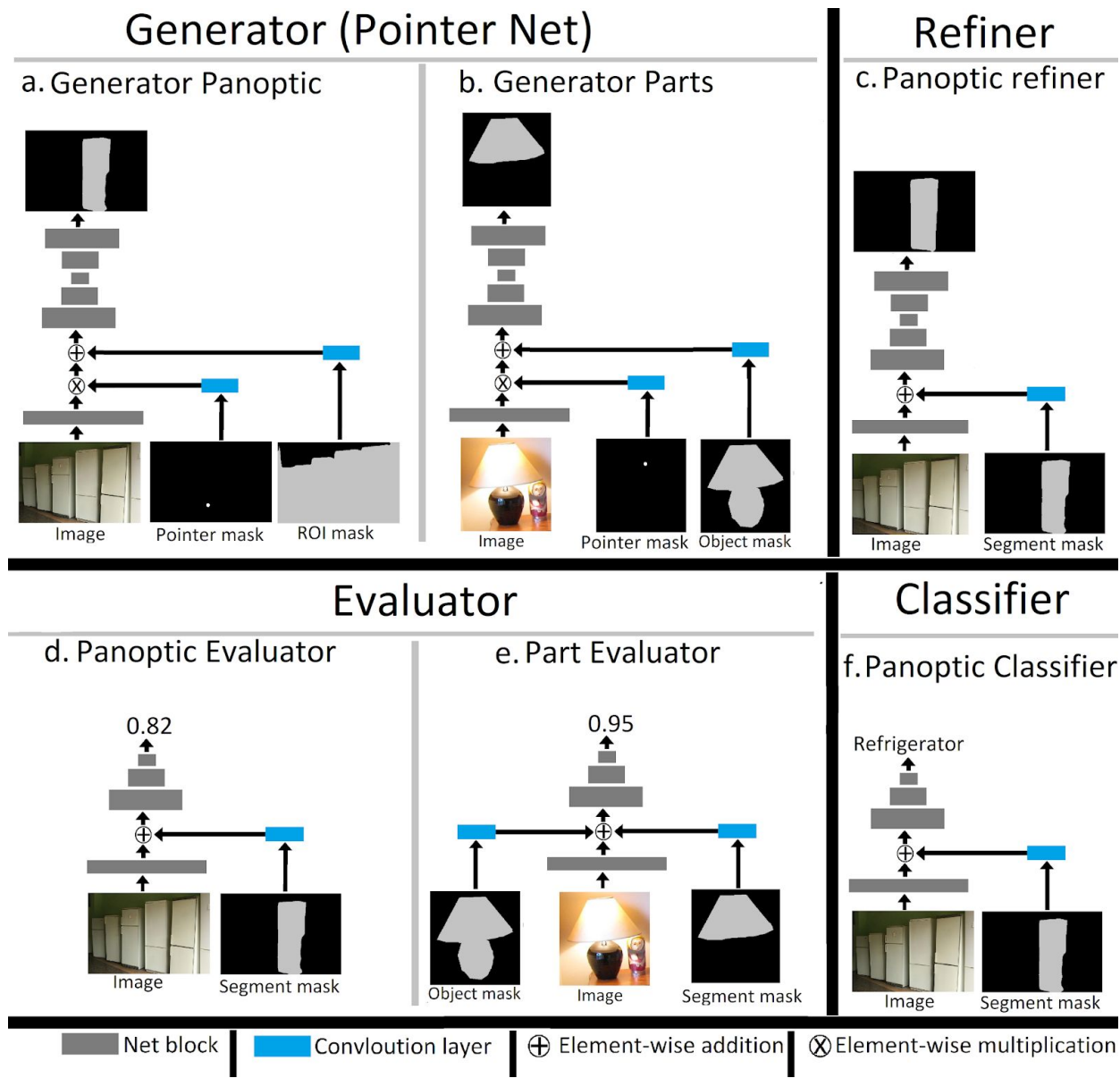

Figure 5: Structure of the nets used in the system. a,b) Generator (Pointer net) receives an image, a pointer point, and an Object/ROI mask and outputs the region of the segment containing the pointer point (within the input mask). c) The refiner net receives an imperfect segment mask and outputs an improved segment mask. d,e) The Evaluator net receives a segment mask and predicts the correspondence (IOU) between the input segment and a real segment in the image. The Parts Evaluator receives the object mask as additional input. f) The segment Classifier receives a segment mask and outputs the segment category.

## 3.3. On the importance of modularity

For generator-evaluator systems, the ability to generate and evaluate each segment separately is essential. To illustrate this, consider the problem of trying to guess a specific string of letters [47]. Assume a string of 1000 letters where each letter can be one of nine possible symbols. A generator that generates a random 1000 letter string, coupled with an evaluator that checks whether this is the correct string, will need around $9^{1000}$ guesses to find the correct string. However, if the letter at each position of the string is generated and evaluated independently, the generator-evaluator will need only around $9 \times 1000$ tries to guess the correct string [33]. The same is approximately true for the segmentation process. A generator/evaluator system that generates and evaluates the full segmentation map in each step will demand exponentially more guesses compared to a modular system that guesses and evaluates each segment separately.

# 4. Implementation details

### Architecture

All the nets used in this work are based on a standard CNN with ResNet architecture (Figure 5) [48]. The pointer and refinement net are both based on standard FCN (Figure 5) [6] with an intermediate pyramid scene parsing (PSP) layer [49] and three upsampling and skip connections layers (Figure 5) [50]. The evaluator and classification nets (Figure 5) are based on the standard ResNet image classification model (Figure 5) [48]. For the evaluator net, the final layer was changed into a single channel prediction, which corresponds to the predicted IOU value (Figure 5). Adding the segment mask (or ROI/Object mask) as an additional input to the nets [8, 9] was done by taking this mask (as a binary 0/1 image) and processing it using a single convolutional layer (Figure 5). The output of this process was merged with the feature map of the ResNet first layer using element-wise addition (Figure 5). Similarly, the pointer point input for the pointer net was introduced by representing the point as a binary mask in the size of the image, where the value of the cell in the pointer point location is one and the rest of the cells are zero (Figure 5). This pointer mask was processed using a single convolutional layer, and the output feature map was merged with the feature map of the ResNet first layer using element-wise multiplication (Figure 5) [8].

### Training for panoptic segmentation

Each of the above nets was trained separately using standard training methods. The training data for the pointer net was created by picking random segments from the annotation of the COCO panoptic training dataset [4]. Segments of things (people, cars) were taken as the full object instance mask. Segments of stuff (sky, grass) were taken as the connected component region of pixels with the same class. Pointer points input for the pointer net were picked by selecting random points within the selected segments. ROI mask input for the pointer net was generated by picking a random segment from the annotation map and using their combined region as the ROI mask. For 50% of the pointer net training iterations, an ROI mask that covers the full image area was used. Training data for the refinement, evaluation and classification nets were generated by running various versions of pointer nets on the COCO panoptic training set. The output of this process was used along with the ground truth annotation as the training data for the refinement, evaluation and classification nets.

For 50% of the training iterations, segments were picked randomly with equal probability for all classes. For the remaining 50% of the training iterations, segments were picked randomly with an equal probability per segment regardless of class (common classes were picked more than rare ones). The training loss for the pointer, classification and refinement nets was the standard cross-entropy. The loss for the evaluator net was the square difference between the predicted and the real IOU of the input segment and closest match segment in the ground truth annotation. Each of the nets was trained on a single TITAN XP GPU for about 1–2 million interactions. The full system composed of the four networks runs in half-precision on a single TITAN XP.

## Training for part segmentation

Training of the system for the part segmentation was done using the ADE20K[40] and Pascal Parts[39] datasets. For each training iteration, several object/stuff masks were chosen randomly (from the GT annotation) with equal probability per object/stuff class. A random part mask was chosen inside each object mask, and a random pointer point was chosen within the part. The net was trained with the image, object mask, and pointer point mask as input (Figure 5). The cross-entropy between the predicted mask and the selected GT part mask was used as the loss. The evaluator was trained with the image, predicted part mask, and object mask as input. The loss was the square difference between the net prediction (as a single number) and the real IOU between the predicted part mask and the GT part mask. The objects segments for each iteration were chosen randomly with an equal probability per class. In addition, 20% of the object classes were not used in the training stage and were used only for the testing of the net. The selection of these unseen classes was made randomly.

## Evaluating parts segmentation

The trained nets were evaluated on the ADE20K[40] and Pascal Parts[39] evaluation sets. The parts regions in the ADE20K are only partly annotated. As a result, it is not possible to evaluate every segment predicted by the net. However, it is possible to evaluate for each GT segment what is the most similar predicted segment in terms of IOU. For each part mask in the GT annotation, the most similar predicted mask was found, and the IOU, precision, and recall, between predicted and GT masks, were used. The average for all classes is given in Table 1. The accuracy for classes that appeared in the training set (familiar) and for classes that did not occur in the training set (unfamiliar) was measured separately (Table 1). The evaluation was limited for part masks with more than 100 pixels and which occupy more than 1% of the object mask.

## COCO dataset and the panoptic quality (*PQ*)

The COCO panoptic dataset [4] is the largest and one of the most general segmentation datasets to date. This dataset contains a large diversity of scenes and about 133 categories including object instances (things) and non-object (stuff). The recognition accuracy is measured by the *panoptic quality* (*PQ*) [4]. *PQ* consists of a combination of *recognition quality* (*RQ*) and *segmentation quality* (*SQ*). For objects, a segment is defined as the region of each individual object instance in the image. For non-objects (stuff), a segment is defined as all pixels in the image in the same semantic category. *RQ* is used to measure the detection accuracy and is given by $RQ = \frac{TP}{TP+(FP+FN)\times 0.5}$, where *TP* (true positive) is the number of predicted segments that match a ground truth segment. *FN* (false negative) is the number of segments in the ground truth annotation that does not match any of the predicted segments. *FP* (false positive) is the number of predicted segments with no matched segment in the ground truth annotation. Matching is defined as an IOU of 50% or more between predicted and ground truth segments of the same category. Segmentation quality (*SQ*) is simply the average IOU of matching segments. *PQ* is calculated as $PQ = RQ \times SQ$. This is calculated for each category separately and then averaged to give the final score.

## 5. Results

Parts segmentation

The results for segmenting objects into parts are given in Table 1 and Figure 6. It can be seen that the GES net has achieved good accuracy for segmenting parts of unfamiliar classes that did not appear in the training set compared to familiar classes that appeared in the training set. This implies that the method managed to generalize to classes it never encountered. Note that while the object of the unseen classes did not appear in the training set, their parts sometimes did. For example, while the airplane class did not appear in the training set, window part class, which is part of an airplane, appears as part of the building that appears in the training set. The generator net was also run without the evaluator, and the generated segments were added in random order. This results in a significantly lower accuracy (Table 1) compared to the generator evaluator system. It can be seen in Table 1 that for the Pascal Parts, the accuracy for unfamiliar classes is higher than for familiar classes. This can be explained by the difference in the segmentation difficulty of different classes and the relatively small number of randomly chosen classes in the unfamiliar set.

| Method | Data-set | With Evaluator | | | Without Evaluator | | | Number of Classes |
| --- | --- | --- | --- | --- | --- | --- | --- | --- |
| | | IOU | Precision | Recall | IOU | Precision | Recall | |
| **Unfamiliar Classes** | ADE20K | 59 | 73 | 78 | 55 | 73 | 73 | 38 |
| **Familiar Classes** | ADE20K | 58 | 76 | 75 | 55 | 73 | 73 | 79 |
| **Unfamiliar Classes** | Pascal | 48 | 60 | 76 | 40 | 65 | 62 | 4 |
| **Familiar Classes** | Pascal | 39 | 54 | 67 | 35 | 66 | 58 | 16 |

**Table 1: Results of the Generator evaluator system for part segmentation. For familiar classes that appear in the training set and unfamiliar classes that did not occur in the training set. Results for the ADE20K and Pascal Parts evaluation sets, averaged for all classes.**

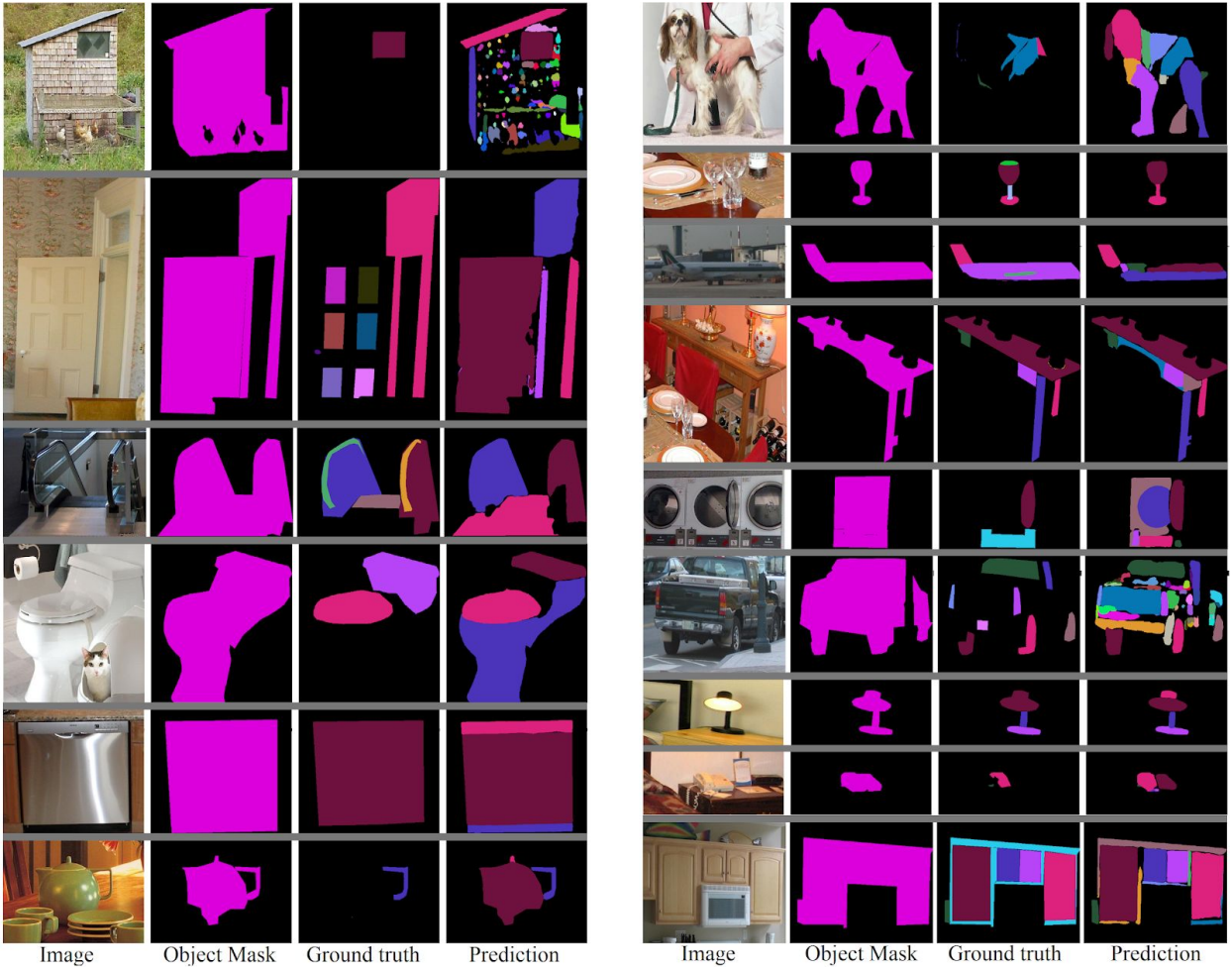

Figure 6: Result of the generator evaluator part segmentation on the ADE20K. For object classes that did not appear in the training stage. Black implies an unsegmented region.

Figure 7: Example results of each of the nets. The ROI mask and the pointer point are the inputs for the pointer net. The predicted segment is the output of the pointer net and the input for the evaluation and refinement nets. The refined segment is the output of the refinement net and input for the classification net. The IOU and class were predicted by the evaluation and classification nets.

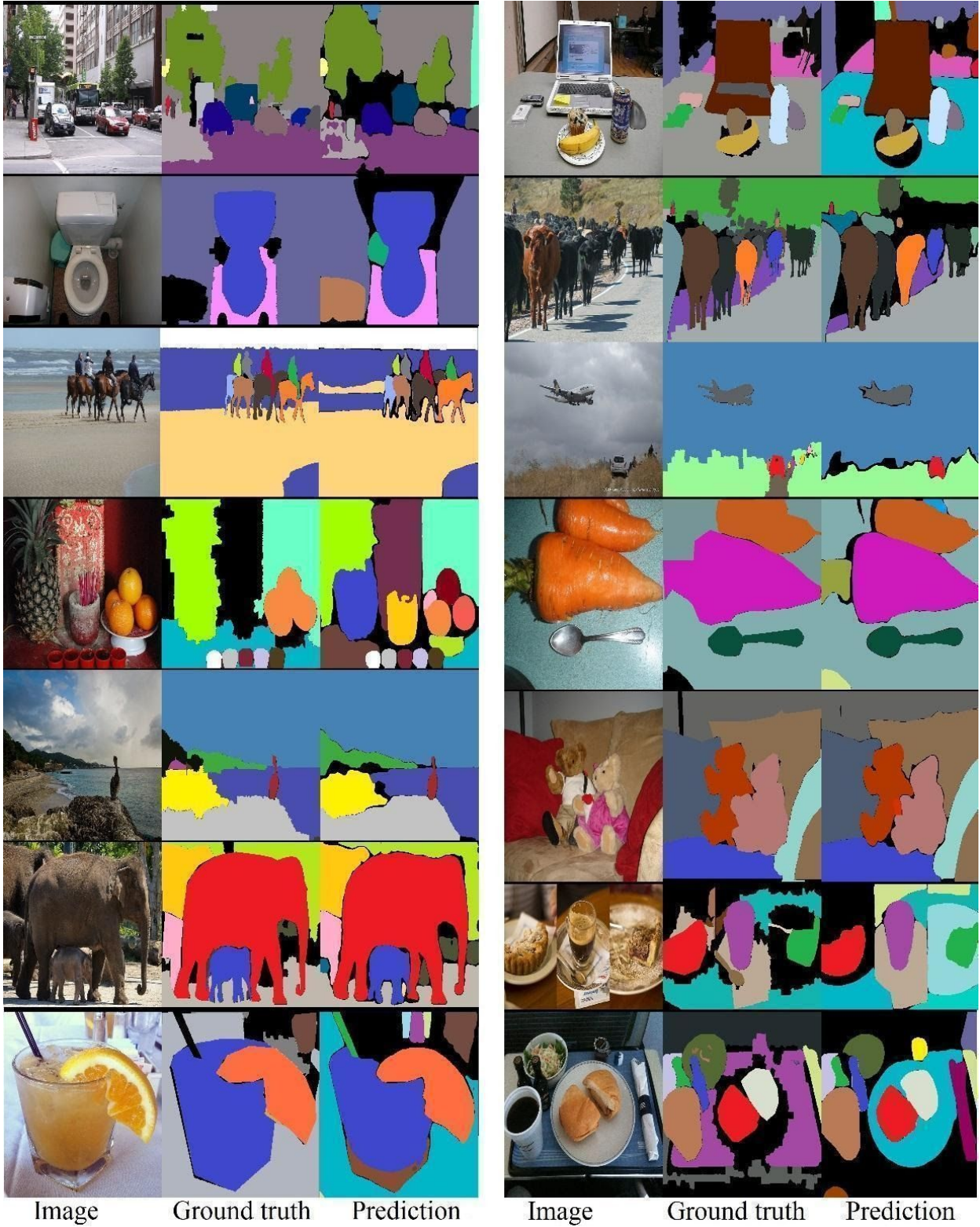

Figure 8: Sample results of the full generator/selector system on the COCO panoptic evaluation set. Black implies unsegmented regions.

| Method | Set | PQ | RQ | SQ | PQ$^{St}$ | RQ$^{St}$ | SQ$^{St}$ | PQ$^{Th}$ | RQ$^{Th}$ | SQ$^{Th}$ |
|---|---|---|---|---|---|---|---|---|---|---|
| **Full system** | Test | **33.7** | **41.4** | **79.6** | **31.5** | **39.3** | **78.4** | **35.1** | **42.9** | **80.4** |
| Full system | Eval | 33.2 | 40.9 | 79.2 | 30.5 | 38.0 | 78.1 | 34.9 | 42.8 | 80.0 |
| No evaluator | Eval | 24.5 | 30.8 | 76.7 | 27.0 | 34.0 | 76.6 | 22.8 | 28.7 | 76.7 |
| Perfect Evaluator | Eval | 39.3 | 48.8 | 78.3 | 35.3 | 44.8 | 77.2 | 41.9 | 51.5 | 79.0 |
| No refinement | Eval | 31.7 | 39.7 | 78.2 | 28.9 | 36.4 | 77.6 | 33.6 | 41.8 | 78.6 |
| Perfect classification | Eval | 45.9 | 56.6 | 79.8 | 54.1 | 66.9 | 80.0 | 40.4 | 49.9 | 79.7 |

Table 2: Results on the COCO panoptic test and evaluation sets. Superscripts Th and St stand for things and stuff (object and not objects). Full system is the main method (Figure 4). No evaluator: all segments were approved without the evaluation stage. Perfect evaluator: the IOU of the evaluator net was replaced by ground truth IOU. No refinement: no refinement stage was used (Figure 4). Perfect classification: the segment class was taken from the ground truth annotation instead of the classifier prediction.

Panoptic segmentation

The results of the panoptic system are given in Figures 7–8 (samples) and Table 2 (statistics). The system (Figure 4) was evaluated on the COCO panoptic test set, resulting in a *PQ* score of 33.7 (Table 2). This value is higher than a few basic implementations of the combined Mask R-CNN and semantic segmentation approach [29]. However, it is significantly lower than the more sophisticated implementation of this method which involves attention [24] and occlusion modules [26, 30].

Effect of the evaluator.

To examine the effect of the evaluator, the system was run with no evaluator (all segments were approved). The result is a significant decrease of 9 points in the *PQ* score (*PQ*=24.5, Table 2). This confirms the importance of the evaluator/selector module. The PQ drop is larger for things then for stuff categories, implying that the evaluator is more important for object classes. To examine the maximum effect of the evaluator, the system was run with a perfect evaluator. The perfect evaluator was simulated by replacing the IOU score predicted by the evaluator with the real IOU extracted from the ground truth annotation. This increased the *PQ* score by 6 points (Table 2), suggesting that mistakes in segments ranking is a moderate source of errors.

Effect of refinement.

To examine the contribution of the refinement net, the system was run without the refinement stage. This results in a drop of 1.5 points in the *PQ* score (Table 2). This implies small but real contributions of the refinement module.

Effect of the classification module.

In order to examine the effect of misclassification, the category generated by the classification net was replaced by the segment real class (taken from the ground truth annotations). The result is a significant increase of 12.7 points in the *PQ* score (Table 2). This implies that misclassification is a major source of errors in the system. This *PQ* increase is particularly large for stuff categories implying misclassification is a bigger problem for none-object classes.

# 6. Conclusion

This work suggests a simple guess and check approach for image segmentation. In this method, one net guesses segments corresponding to objects and regions in the image, while a second net checks and selects the best segments. This system was combined with two additional nets, one for segment refinement and the other for segment classification. This creates a single modular system capable of panoptic and parts segmentation. One advantage of this modular approach is that each module can be kept as simple and general as possible [51, 52]. In addition, each module could be replaced or used independently without the need for any retraining. For example, removing the classification module will turn the system into a class agnostic segmentation system capable of segmenting regions of unknown categories that were not a part of the training set. The ability of this system to work on specific image regions and unfamiliar classes enables it to solve problems like class agnostic part segmentation, which involves splitting objects of classes it did not encounter in training into parts. An ability that, as far as we know, has not been demonstrated by any other neural net. In addition, the system was evaluated on the COCO panoptic dataset, while the results are significantly lower than top methods based on mask R-CNN and Semantic segmentation, they are at the same level as the basic implementation of these methods.

**Acknowledgments:** We thank Natural Resources Canada for their generous support for this project. We also thank Anders G. Froseth for his support of our work on artificial intelligence. We thank the Vector Institute for computational resources. We also thank the Office of Naval Research for support.

# Supporting material

*Code for the part segmentation has been made available at:*
*https://github.com/sagieppel/Splitting-unfamiliar-objects-and-stuff-in-images-into-parts-using-neural-nets*

*Code for the Panoptic GES Net has been made available at:*
https://github.com/sagieppel/Generator-evaluator-selector-net-a-modular-approach-for-panoptic-segmentation